\documentclass[letterpaper]{article}
\usepackage[ansinew]{inputenc}
\usepackage{times}  %Required
\usepackage{helvet}  %Required
\usepackage{courier}  %Required
\usepackage{url}  %Required
\usepackage{graphicx}  %Required
\usepackage{amsfonts}
\usepackage{amsmath,amssymb}
\usepackage{algorithm}
\usepackage{algorithmicx}
\usepackage{algpseudocode}
\usepackage{subcaption}
\usepackage{multirow}
\usepackage{authblk}

\newtheorem{theorem}{Theorem}

\DeclareMathOperator*{\E}{\mathbb{E}}
\DeclareMathOperator*{\argmax}{arg\,max}
\newcommand{\rpm}{\sbox0{$1$}\sbox2{$\scriptstyle\pm$}
  \raise\dimexpr(\ht0-\ht2)/2\relax\box2 }

\providecommand{\keywords}[1]
{
  \small	
  \textbf{\textit{Keywords---}} #1
}

\title{Effective Exploration for Deep Reinforcement Learning via Bootstrapped Q-Ensembles under Tsallis Entropy Regularization}

\author[1]{Gang Chen\thanks{aaron.chen@ecs.vuw.ac.nz}}
\author[1]{Yiming Peng\thanks{yiming.peng@ecs.vuw.ac.nz}}
\author[1]{Mengjie Zhang\thanks{mengjie.zhang@ecs.vuw.ac.nz}}
\affil[1]{School of Engineering and Computer Science,

Victoria University of Wellington, New Zealand}

\begin{document}

\maketitle

\begin{abstract}
Recently \emph{deep reinforcement learning} (DRL) has achieved outstanding success on solving many difficult and large-scale RL problems. However the high sample cost required for effective learning often makes DRL unaffordable in resource-limited applications. With the aim of improving sample efficiency and learning performance, we will develop a new DRL algorithm in this paper that seamless integrates \emph{entropy-induced} and \emph{bootstrap-induced} techniques for efficient and deep exploration of the learning environment. Specifically, a general form of \emph{Tsallis entropy regularizer} will be utilized to drive entropy-induced exploration based on efficient approximation of optimal action-selection policies. Different from many existing works that rely on action dithering strategies for exploration, our algorithm is efficient in exploring actions with clear exploration value. Meanwhile, by employing an ensemble of Q-networks under varied Tsallis entropy regularization, the diversity of the ensemble can be further enhanced to enable effective bootstrap-induced exploration. Experiments on Atari game playing tasks clearly demonstrate that our new algorithm can achieve more efficient and effective exploration for DRL, in comparison to recently proposed exploration methods including Bootstrapped Deep Q-Network and UCB Q-Ensemble.
\end{abstract}


\section{Introduction}
\label{sec-intro}

In recent years, \emph{deep reinforcement learning} (DRL) has been extensively and successfully utilized by computer systems to autonomously learn to solve many challenging problems such as robotics control \cite{lillicrap2015,schulman2015}, video game playing \cite{mnih2015,van2016}, and road traffic management \cite{li2016}. However, in order to achieve its learning goals, an RL agent must often use a huge amount of sampled data to train its \emph{deep neural networks} (DNNs). Since data sampling is realized through direct trial-and-error interactions with the learning environment, the high \emph{sample cost} usually makes DRL unaffordable in resource-limited applications \cite{wu2017,wang2016,chen20181}.

In order to improve sample efficiency, an RL agent must carefully manage its exploration of the learning environment. Osband \emph{et al.} recently proposed the idea of ``\emph{deep exploration}" to emphasize on the requirement for the agent to learn effectively within a reasonable time frame by considering not only the immediate benefits of taking any action but also the long-term impact of the action on future learning, thereby properly synthesizing efficient exploration with effective generalization \cite{osband2016,osband2014,osband2017}. Guided by this requirement, Bootstrapped Deep Q-Network (Bootstrapped DQN) has been proposed lately to drive deep and efficient exploration \cite{osband20161}.

Bootstrapped DQN was inspired by the posterior sampling method for RL with near-optimal regret bounds \cite{osband2013}. However, instead of sampling and solving numerous \emph{Markov Decision Processes} (MDPs), Bootstrapped DQN approximates a posterior model over optimal Q-functions (also known as the state-action value functions) at much affordable computation cost. This is shown to easily outperform \emph{action dithering strategies} for exploration such as $\epsilon$-greedy or softmax action sampling techniques \cite{kakade2003,strehl2007}. For this purpose, an ensemble of randomly initialized Q-networks (or Q-functions) will be maintained consistently during RL. Empirical results showed that effective deep exploration can be achieved in practice by randomly choosing one of the Q-networks to guide multi-step interactions with the learning environment.

Besides Bootstrapped DQN, the UCB Q-Ensemble method proposed in \cite{chen2017ucb} also relies on learning concurrently an ensemble of Q-networks. However it adopts an approximated \emph{upper-confidence bound} over Q-values produced by these Q-networks to steer exploration. Although highly competitive performance has been witnessed on Atari game playing tasks, theoretical studies suggest that precise calculation of such confidence bounds can be computationally intractable \cite{russo2013,russo2014}.

Similar to Bootstrapped DQN and UCB Q-Ensemble, we employ an ensemble of Q-networks to achieve deep exploration. However, without relying on actions with either the highest Q-values or upper-confidence bounds for exploration, we generalize action selection by studying policies under entropy regularization. This generalization enables us to develop a new form of optimal stochastic policies, thereby relieving the dependency on randomly initialized Q-networks as the main source of randomness for deep exploration \cite{chen2017ucb,osband20161}.

In the literature, Shannon entropy is frequently utilized to regularize action selection, giving rise to optimal policies that exhibit \emph{softmax action-selection behaviors} \cite{donoghue2017,nachum2017,haarnoja2017,schulman2017e}. While softmax distributions naturally bring stochasticity to deep exploration, they are prone to assigning non-negligible probability mass to actions with negligible exploration value \cite{lee2017,nachum2018}.

Tsallis entropy is an important extension of Shannon entropy \cite{tsallis1993,tsallis1994}. A special case of \emph{Tsallis entropy} has been studied in \cite{lee2017} to tackle sparse MDP problems. When applied to DRL, Tsallis entropy allows an RL agent to concentrate on exploring actions that deserve further exploration. Due to this reason, \emph{Tsallis entropy regularization is deemed a key mechanism for efficient deep exploration} in this paper. Moreover, without being restricted to any specific setting of Tsallis entropy as in \cite{lee2017,nachum2018}, a general form of Tsallis entropy will be studied the first time in literature to guide DRL.

Based on computationally efficient approximation of optimal policies under general Tsallis entropy regularization, a new deep exploration algorithm involving an ensemble of deep Q-networks will be further developed in this paper. The newly proposed algorithm will be called the \emph{Bootstrapped Q-Ensemble under Tsallis Entropy Regularization} (BQETR) algorithm. Each Q-network in BQETR adopts a different setting of the Tsallis entropy regularizer in order to achieve high ensemble diveristy. Meanwhile, the \emph{regularization coefficient} is kept the same for all Q-networks and will be gradually reduced to 0 as an RL agent gains increasingly more experience from its learning environment. In this way we expect to seamlessly integrate \emph{entropy-induced exploration} with \emph{bootstrap-induced exploration} for effective RL.

Empirical studies have been performed on benchmark Atari game playing tasks. Our experiment results clearly show that BQETR has achieved significantly better sample efficiency and performance than Bootstrapped DQN and UCB Q-Ensemble. We therefore believe that BQETR is an effective and efficient method for deep exploration and RL.

\iffalse
The rest of this paper is organized as follows. Section \ref{sec-relatedwork} reviews closely related works found in the recent literature. A new algorithm for deep Q-learning with Tsallis entropy regularization will be developed in Section \ref{sec-regq}. Based on this algorithm, a new approach for deep exploration driven by bootstrapped Q-ensembles will be further proposed in Section \ref{sec-deepexplor}, leading to the development of BQETR. Experiment study of BQETR will be conducted and reported in Section \ref{sec-exp}. Finally Section \ref{sec-con} will conclude this paper and briefly discuss potential future works.
\fi

\section{Related Works}
\label{sec-relatedwork}

Huge efforts have been devoted to developing efficient and effective exploration strategies for RL. Particularly, provably efficient exploration techniques have been studied based on the idea of Bayes optimal policies \cite{ghavamzadeh2015} and clearly revealed the importance of multi-step exploration \cite{kearns2002}. Further studies along this line also demonstrated the inefficiency of $\epsilon$-greedy and softmax exploration techniques on large RL problems \cite{brafman2002,strehl2006,auer2007,dann2015}. In view of the fact that many existing DRL algorithms rely on such simple methods for exploration \cite{mnih2015,van2016}, developing new exploration methods for effective DRL has great value both in theory and in practice.

Among all the exploration techniques proposed so far, a notable series of research works clearly highlighted the advantages of exploration through randomized value functions \cite{osband2013,osband2014,osband2016,osband2017}. Specifically, the randomized least-squares value iteration (RLSVI) algorithms proposed in \cite{osband2017} extended traditional least-squares value iteration methods through randomly sampling \emph{statistically plausible value functions}. However, efficient sampling often requires value functions to be linear with respect to their parameters and may not be suitable for DRL \cite{osband2014,osband2017}. To cope with this issue, Bootstrapped DQN has been developed recently in \cite{osband20161} to approximately sample value functions modeled as DNNs.

Besides bootstrapping, to achieve entropy-induced exploration, the training of action-selection policies is often reshaped by a Shannon entropy regularizer, resulting in various soft-Q style algorithms for DRL \cite{nachum2017,donoghue2017,haarnoja2017}. For example, Nachum \emph{et al.} developed an off-policy algorithm based on a multi-step consistency equation for entropy-regularized RL \cite{nachum2017}. Haarnoja \emph{et al.} conducted research on soft-Q learning in high-dimensional action spaces \cite{haarnoja2017}. In \cite{donoghue2017,schulman2017e}, policy gradient training is shown as equivalent to soft-Q learning. This insightful understanding enables an RL agent to combine policy gradient with Q-learning for effective sample reuse \cite{donoghue2017}. Different from these research works, efficient entropy-induced exploration is realized in this paper through Tsallis entropy regularization \cite{lee2017,nachum2018}.

This paper is similar to \cite{lee2017,nachum2018} since they all leverage on Tsallis entropy to regularize policy optimization. However, different from these research works that studied only a specific setting of Tsallis entropy, we will consider general forms of Tsallis entropy so as to maintain strong diversity in a Q-ensemble. It also allows us to treat Shannon entropy regularization as a special case of our research. Moreover, the integrated use of Tsallis entropy and bootstrapping mechanism for deep exploration further separates this paper apart from most of the previous works.

\section{Entropy Regularized Q-Learning}
\label{sec-regq}

In this section, we will introduce the RL problem first, followed by a quick review of Q-learning and policy gradient learning techniques. Afterwards, a new deep Q-learning algorithm under Tsallis entropy regularization will be developed. The \emph{Bellman residue} of the newly proposed algorithm will also be analyzed under an extreme circumstance.

\subsection{The Reinforcement Learning Problem}
\label{sub-sec-rl}

This paper studies general RL problems that can be described by an MDP with an arbitrary set of states $s\in\mathbb{S}$ and a finite set of actions $a\in\mathbb{A}$ \cite{donoghue2017}. Such problems appear frequently in literature including robotics control and video game playing \cite{mnih2015,van2016,chen2018}. At each time step $t$, an RL agent observes its environment and determines its current state $s_t$. It subsequently selects and performs an action $a_t$, driving the environment to move to its next state $s_{t+1}$ with a probability $P(s_t,a_t,s_{t+1})$ which is unknown to the agent. Meanwhile, a scalar reward $r(s_t,a_t)$ is provided as the immediate feedback of performing action $a_t$. Starting from any initial state $s_0$, the agent is required to perform a long (sometimes infinite) sequence of actions in order to obtain the \emph{discounted total return} defined below
\begin{equation}
J(\pi)=\E_{s_0,\pi} \left( \sum_{t=0}^{\infty} \gamma^t r_t \right)
\label{equ-j}
\end{equation}

\noindent
where the expectation in \eqref{equ-j} is conditional on initial state $s_0$ and $\pi$. Here $\pi$ refers to a \emph{stochastic action-selection policy} that the RL agent follows to determine the probability $\pi(s_t,a)$ of performing any action $a$ in any state $s_t$. Obviously $\sum_{a\in\mathbb{A}} \pi(s_t,a)=1$ is an important constraint for $\pi$ to be well-defined. Moreover, $\gamma$ takes its value in $[0,1)$ and serves as a discount factor for the RHS of \eqref{equ-j} to be meaningful. With an MDP described as above, the ultimate goal of RL is hence to identify the \emph{optimal policy} $\pi^*$ that maximizes $J(\pi^*)$, i.e.
\begin{equation}
\pi^*=\argmax_{\pi} J(\pi)
\end{equation}

\noindent
In an effort to learn $\pi^*$, an RL agent may choose to first learn the Q-function with respect to some non-optimal policy $\pi$, as defined below
\begin{equation}
Q^{\pi}(s_t,a_t)=\E_{\pi,s_t,a_t} \left( \sum_{k=0}^{\infty} \gamma^k r_{k+t} \right)
\label{equ-q-func}
\end{equation}

\noindent
Given $Q^{\pi}$ in \eqref{equ-q-func}, the value of state $s_t$ under policy $\pi$ is determined further as
\begin{equation}
V^{\pi}(s_t)=\E_{s_t,\pi} \left( \sum_{k=0}^{\infty} \gamma^k r_{k+t} \right)=\sum_{a\in\mathbb{A}}\pi(s_t,a) Q^{\pi}(s_t,a)
\end{equation}

\noindent
To ease discussion, we denote the value of any state $s$ under the optimal policy $\pi^*$ as $V^*(s)$. Accordingly $Q^*(s,a)$ represents the maximum Q-value achievable as a result of performing action $a$ in state $s$.

\subsection{Q-Learning and Policy Gradient}
\label{sub-sec-pgq}

In value function based methods for DRL, the Q-function (or V-function) is represented as a DNN with numerous parameters. Through updating these parameters, we can bring the Q-value outputs from such deep Q-networks as close to the fixed point of the \emph{Bellman equation} as possible. Two versions of the Bellman equation are typically studied in the literature. Each version is associated with a different \emph{Bellman operator}, i.e. $\mathcal{T}^{\pi}$ and $\mathcal{T}^*$, as defined below.
\begin{equation}
\begin{split}
\mathcal{T}^{\pi} Q^{\pi}(s,a)= & r(s,a)+\\
 &\gamma \sum_{s'} P(s,a,s')\sum_{b\in\mathbb{A}} \pi(s',b) Q^{\pi}(s',b)
\end{split}
\label{equ-tpi}
\end{equation}
\begin{equation}
\mathcal{T}^* Q^*(s,a)=r(s,a)+\gamma \sum_{s'} P(s,a,s')\max_{b\in\mathbb{A}} Q^*(s',b)
\label{equ-t*}
\end{equation}

\noindent
Clearly $\mathcal{T}^{\pi}$ in \eqref{equ-tpi} applies to $Q^{\pi}$ which is the fixed point of the Bellman equation $\mathcal{T}^{\pi}Q(s,a)=Q(s,a)$. Likewise $\mathcal{T}^*$ in \eqref{equ-t*} applies to $Q^*$ which is the fixed point of the Bellman equation $\mathcal{T}^*Q(s,a)=Q(s,a)$. It is well-known that both $\mathcal{T}^{\pi}$ and $\mathcal{T}^*$ are $\gamma$-contraction mappings in the sup-norm and are suitable to drive value function learning \cite{bertsekas1995}. Specifically, in DQN \cite{mnih2015}, an approximation of $\mathcal{T}^*$ based on a batch of previously sampled state transition data $\mathcal{B}=\{ (s,a,s',r(s,a)) \}$ is utilized to update the Q-network parameters in the direction of minimizing the loss function below
\begin{equation}
L^*(\theta)=\sum_{(s,a,s',r)\in\mathcal{B}} \left( Q_{\theta}(s,a) - \tilde{\mathcal{T}}^*(s,a,s',r) \right)^2
\label{equ-loss-dqn}
\end{equation}

To avoid overestimation in the original design of DQN, Double DQN is typically used in practice by employing a separate target Q-network to calculate $\tilde{\mathcal{T}}^*$ in \eqref{equ-loss-dqn} \cite{van2016}. The equation below shows more details.
\begin{equation}
\tilde{\mathcal{T}}^*(s,a,s',r)=r(s,a)+\gamma Q_{\theta^-}\left(
s',\argmax_{b\in\mathbb{A}} Q_{\theta}(s',b)
\right)
\label{equ-t*t}
\end{equation}

\noindent
where in \eqref{equ-loss-dqn} and \eqref{equ-t*t}, $Q_{\theta}$ refers to the Q-network parameterized by $\theta$ and $Q_{\theta^-}$ stands for the target Q-network parametrized by $\theta^-$. Clearly by minimizing $L^*$ in \eqref{equ-loss-dqn}, $Q_{\theta}$ has the aim to approximate $Q^*$. Similar loss function has also been defined for $Q_{\theta}$ to precisely estimate $Q^{\pi}$ with respect to arbitrary policy $\pi$.

Under the general actor-critic framework for RL, assume that a stochastic policy $\pi$ is implemented as a DNN parameterized by $\omega$. According to the policy gradient theorem \cite{sutton2000}, $\omega$ should be updated in the direction of
\begin{equation}
\frac{\partial J(\pi)}{\partial \omega}=\E_{(s,a)\sim\pi} \left( Q^{\pi}(s,a) \frac{\partial \log\pi(s,a)}{\partial \omega} \right)
\label{equ-pg-the}
\end{equation}

\noindent
Here the expectation is taken over all possible state-action pairs $(s,a)$ with probability $d^{\pi}(s)\pi(s,a)$ and $d^{\pi}(s)$ gives the \emph{discounted distribution of states} defined in \cite{sutton2000}.

\begin{algorithm}[!ht]
 \begin{algorithmic}[1]
 \item {\bf Input}: a Q-network, $q$ value for the Tsallis entropy regularizer, $\alpha_0$ for the initial regularization coefficient, and a replay buffer $\mathcal{B}$ that stores past state-transition samples for training
 \item {\bf for} each problem episode {\bf do}:
 \item \ \ \ \ Obtain initial state $s_0$ from environment
 \item \ \ \ \ {\bf for} $t=1,\ldots$ until end of episode {\bf do}:
 \item \ \ \ \ \ \ \ \ Sample action $a_t$ according to \eqref{equ-psa-prop1}
 \item \ \ \ \ \ \ \ \ Perform $a_t$
 \item \ \ \ \ \ \ \ \ Add $(s_t,a_t,s_{t+1},r_t)$ to $\mathcal{B}$
 \item \ \ \ \ \ \ \ \ {\bf if} learning interval is reached {\bf do}:
 \item \ \ \ \ \ \ \ \ \ \ \ \ Sample mini-batch from $\mathcal{B}$
 \item \ \ \ \ \ \ \ \ \ \ \ \ Update Q-network to minimize $L^{\pi^*_{\alpha}}(\theta)$ in

 \ \ \ \ \ the mini-batch
 \item \ \ \ \ \ \ \ \ \ \ \ \ Reduce $\alpha$ linearly by $\Delta_{\alpha}$ until 0
 \end{algorithmic}
\caption{An Algorithm for Deep Q-Learning under Tsallis Entropy Regularization}
\label{alg-1}
\end{algorithm}

\subsection{Q-Learning under Tsallis Entropy Regularization}
\label{sub-sec-qte}

While training policy networks based on \eqref{equ-pg-the}, in order to prevent a policy from converging too fast and therefore leaving no opportunity for future exploration, it is a common practice to introduce an extra \emph{entropy regularizer}. As a consequence, the policy network parameters $\omega$ can be updated according to
\begin{equation}
\Delta\omega \propto \E_{(s,a)\sim\pi} \left( Q^{\pi}_{\theta}(s,a) \frac{\partial log \pi(s,a)}{\partial \omega} + \alpha \frac{\partial H^{\pi}(s)}{\partial \omega} \right)
\label{equ-reg-pg}
\end{equation}

\noindent
where $H^{\pi}$ denotes the entropy of policy $\pi$ and $\alpha>0$ is the \emph{entropy regularization coefficient}. Previously, Shannon entropy as defined below is frequently utilized for regularized policy training.
\begin{equation}
H^{\pi}_I (s)=-\sum_{a\in\mathbb{A}} \pi(s,a)\log\pi(s,a)
\label{equ-it-entr}
\end{equation}

\noindent
Subject to entropy regularization in \eqref{equ-it-entr} with coefficient $\alpha$ in \eqref{equ-reg-pg}, it can be shown that the optimal policy $\pi^*_{\alpha}$ and the corresponding optimal Q-function $Q^{\pi^*_{\alpha}}$ obey the following equation \cite{schulman2017e},
\begin{equation}
\pi^*_{\alpha}(s,a)=\frac{ \exp(Q^{\pi^*_{\alpha}}(s,a)/\alpha) }{ \sum_{b\in\mathbb{A}} \exp(Q^{\pi^*_{\alpha}}(s,b)/\alpha) }
\label{equ-soft-q}
\end{equation}

In this paper, instead of using the softmax distribution in \eqref{equ-soft-q} to guide entropy-induced exploration which can exhibit poor efficiency in practice \cite{osband2017}, we decide to study Tsallis entropy based regularizer as defined below.
\begin{equation}
H^{\pi}_q (s)=\frac{1}{q-1} \left( 1-\sum_{a\in\mathbb{A}} \pi(s,a)^q \right)
\label{equ-tsa-entr}
\end{equation}

\noindent
It can be shown that $\lim_{q\rightarrow 1} H^{\pi}_q (s)=H^{\pi}_I (s)$, for any $s\in\mathbb{S}$. We can hence consider soft-Q learning as demonstrated by \eqref{equ-soft-q} as a special case of our new Q-learning method. In \cite{lee2017,nachum2018}, a specific setting of \eqref{equ-tsa-entr} with $q=2$ has been studied to derive a fixed-form representation of the optimal policy. In this paper, on the other hand, we are interested in the general form of Tsallis entropy in \eqref{equ-tsa-entr} with $q>1$.

Analogous to the analysis presented in \cite{donoghue2017}, we can represent the RHS of \eqref{equ-reg-pg} as $f(\omega)$. Meanwhile let $g^{\pi}(s)=\sum_{a\in\mathbb{A}}\pi(s,a)$. Clearly when $\omega$ in \eqref{equ-reg-pg} reaches its fixed point (or optima), no further updating of $\omega$ in the direction of $f(\omega)$ is possible without violating the constraint that $g^{\pi}(s)=1$ for any $s\in\mathbb{S}$. This means that, with the optimal policy parameters $\omega^*$, $f(\omega^*)$ belongs to the span of the vectors $\{\frac{\partial g^{\pi}(s)}{\partial \omega}\}$, i.e.
\begin{equation}
f(\omega^*)=\sum_{s\in\mathbb{S}} \lambda(s) \left. \frac{\partial g^{\pi}(s)}{\partial \omega} \right|_{\omega=\omega^*}
\label{equ-pol-opt}
\end{equation}

\noindent
where for every state $s$, the Lagrange multiplier $\lambda(s)$ in \eqref{equ-pol-opt} ensures that $g^{\pi}(s)=1$. Meanwhile we can determine $\frac{\partial H^{\pi}_q (s)}{\partial \omega}$ as
\begin{equation}
\begin{split}
\frac{\partial H^{\pi}_q (s)}{\partial \omega} &=\frac{q}{q-1} \sum_{a\in\mathbb{A}} \pi(s,a)^{q-1} \frac{\partial \pi(s,a)}{\partial \omega} \\
&=\frac{q}{q-1} \sum_{a\in\mathbb{A}} \pi(s,a)^q \frac{\partial \log\pi(s,a)}{\partial \omega}
\end{split}
\label{equ-H-deri}
\end{equation}

\noindent
By substituting \eqref{equ-H-deri} into \eqref{equ-reg-pg} and also taking into account \eqref{equ-pol-opt}, the optimal condition for policy parameters $\omega^*$ becomes
\begin{equation}
\begin{split}
&\E_{(s,a)\sim\pi} \\
& \left( \left( Q^{\pi}(s,a)-\frac{\alpha q}{q-1}\pi(s,a)^{q-1}-c(s) \right)\frac{\partial \log\pi(s,a)}{\partial \omega} \right) \\
&=0
\end{split}
\label{equ-opt-pp}
\end{equation}

\noindent
where $c(s)$ stands for $\lambda(s)$ in \eqref{equ-pol-opt} adjusted according to the discounted distribution of state $s$. To solve the equation in \eqref{equ-opt-pp}, similar to \cite{donoghue2017}, it is eligible to consider each state $s$ separately. Particularly, in any state $s$ and $\forall a\in\mathbb{A}$, we have
\begin{equation}
Q^{\pi}(s,a)-\frac{\alpha q}{q-1}\pi(s,a)^{q-1}-c(s)=0\ \text{or}\ \frac{\partial \pi(s,a)}{\partial \omega}=0
\label{equ-sa-opt}
\end{equation}

\noindent
It is straightforward to verify the solution below of \eqref{equ-sa-opt},
\begin{equation}
\pi^*_{\alpha}(s,a)=\sqrt[q-1]{ \max\left( \left(\frac{Q^{\pi^*_{\alpha}}(s,a)}{\alpha}-\frac{c(s)}{\alpha}\right), 0\right) \frac{q-1}{q}  }
\label{equ-pi-sol}
\end{equation}

\noindent
with $\pi^*_{\alpha}$ representing the optimal policy of the entropy-regularized policy gradient learning problem described in \eqref{equ-reg-pg}. $Q^{\pi^*_{\alpha}}$ stands for the respective Q-function for policy $\pi^*_{\alpha}$. Meanwhile, $c(s)$ in \eqref{equ-pi-sol} ensures that the condition $g^{\pi}(s)=1$ holds consistently. Notice that for certain action $a$, it is possible for $Q^{\pi^*_{\alpha}}(s,a)-c(s) < 0$. For such an action, the validity of \eqref{equ-pi-sol} is ensured by letting $\pi^*(s,a)=0$, therefore $\frac{\partial \pi^*(s,a)}{\partial \omega}=0$. In other words, only a portion of actions in $\mathbb{A}$ may be explored in any state $s$, thereby encouraging efficient exploration. Particularly, when $q=2$, it can be shown that \cite{lee2017}
\begin{equation}
c(s)=\alpha\frac{\sum_{a\in S(s)} \frac{Q^{\pi^*_{\alpha}}(s,a)}{\alpha} -q }{\| S(s) \|}
\label{equ-cs-opt}
\end{equation}

\noindent
with $S(s)$ representing the set of actions with non-zero chance of exploration in state $s$, as determined below.
\begin{equation}
S(s)=\left\{a_i \left| q+i\frac{ Q^{\pi^*_{\alpha}}(s,a_i) }{\alpha} > \sum_{j=1}^i \frac{ Q^{\pi^*_{\alpha}}(s,a_i) }{\alpha}  \right. \right\}
\label{equ-ss-cond}
\end{equation}

\noindent
where $a_i$ denotes the action with the $i$-th highest Q-value in state $s$. If $q\neq 2$, closed-form representation of $c(s)$ and $S(s)$ may not exist. Therefore, in order to estimate $\pi^*_{\alpha}$, we have developed two efficient approximation techniques in our Q-Learning algorithm. Specifically, because whenever $\pi^*_{\alpha}(s,a)>0$ for any state $s$ and action $a$, $Q^{\pi^*_{\alpha}}(s,a)-c(s)>0$. For such action $a$, we can establish a first order approximation of $\pi^*_{\alpha}$ in \eqref{equ-pi-sol} based on
\begin{equation}
\begin{split}
\pi^*_{\alpha}(s,a)\approx & 1 + \frac{1}{q-1} \left(  \left( \frac{Q^{\pi^*_{\alpha}}(s,a)}{\alpha}-\frac{c(s)}{\alpha} \right)\frac{q-1}{q} -1 \right) \\
& + o\left(\left(\frac{Q^{\pi^*_{\alpha}}(s,a)}{\alpha}-\frac{c(s)}{\alpha}\right) \frac{q-1}{q} - 1\right)
\end{split}
\label{equ-add-1}
\end{equation}

\noindent
Now apply the constraint in \eqref{equ-add-2} over all actions belonging to $S(s)$,
\begin{equation}
\sum_{a\in S(s)} \pi^*_{\alpha}(s,a) = 1
\label{equ-add-2}
\end{equation}

\noindent
we can obtain the result
\begin{equation}
c(s)\approx \alpha\frac{\sum_{a\in S(s)} \frac{Q^{\pi^*_{\alpha}}(s,a)}{\alpha} -q }{\| S(s) \|} + \alpha\left(q-\frac{q}{q-1} \right)
\label{equ-cs-approx}
\end{equation}

\noindent
Clearly, when $q=2$, $c(s)$ as approximated in \eqref{equ-cs-approx} is identical to $c(s)$ in \eqref{equ-cs-opt}. Meanwhile, we can  check the condition of $\frac{Q^{\pi^*_{\alpha}}(s,a)}{\alpha}>\frac{c(s)}{\alpha}$ whenever $a\in S(s)$. Apparently, only actions associated with high Q-values in state $s$ have the chance to be performed by an RL agent. Suppose that $\{a_1,\ldots,a_m\}$ are the actions with the $m$ highest Q-values. For $S(s)$ to contain all these actions, we must make sure that
\begin{equation}
\frac{Q^{\pi^*_{\alpha}}(s,a)}{\alpha} > \frac{ \sum_{i=1}^m \frac{Q^{\pi^*_{\alpha}}(s,a_i)}{\alpha} -q}{m} + \left( q-\frac{q}{q-1} \right)
\label{equ-add-3}
\end{equation}

\noindent
Therefore,
\begin{equation}
m \frac{Q^{\pi^*_{\alpha}}(s,a)}{\alpha} + q > \sum_{i=1}^m \frac{Q^{\pi^*_{\alpha}}(s,a_i)}{\alpha} + m \left( q-\frac{q}{q-1} \right)
\label{equ-add-4}
\end{equation}

\noindent
Based on \eqref{equ-add-4}, $S(s)$ can now be estimated immediately as in \eqref{equ-ss-approx-cond}.
\begin{equation}
S(s)\approx\left\{a_i \left|
\begin{array}{l}
q+i\frac{ Q^{\pi^*_{\alpha}}(s,a_i) }{\alpha} > \\
\sum_{j=1}^i \frac{ Q^{\pi^*_{\alpha}}(s,a_i) }{\alpha} + i \left( q-\frac{q}{q-1} \right)
\end{array}
\right. \right\}
\label{equ-ss-approx-cond}
\end{equation}

Due to the inherent error involved in approximating $c(s)$ and $S(s)$ through \eqref{equ-cs-approx} and \eqref{equ-ss-approx-cond} respectively, we do not know for sure which action can be safely ignored for future exploration. Without missing any potentially valuable actions, the second technique to approximate $\pi^*_{\alpha}$ is to use the \emph{softplus function} \cite{dugas2001} as a smooth implementation of $\max(\cdot,0)$ in \eqref{equ-pi-sol}. Specifically,
\begin{equation}
\pi^*_{\alpha}(s,a)\propto \sqrt[q-1]{\delta\left( \frac{Q^{\pi^*_{\alpha}}(s,a)}{\alpha}-\frac{c(s)}{\alpha} \right)}
\label{equ-psa-prop1}
\end{equation}

\noindent
where the softplus function $\delta$ is defined as $\delta(x)=\log(1+\exp(x))$. Based on \eqref{equ-psa-prop1}, we can learn a Q-network parameterized by $\theta$ with the aim of minimizing the loss function $L^{\pi^*_{\alpha}}(\theta)$. $L^{\pi^*_{\alpha}}(\theta)$ is calculated based on \eqref{equ-loss-dqn} where the Bellman operator $\mathcal{T}^*$ is replaced by $\mathcal{T}^{\pi^*_{\alpha}}$. Driven by this idea, we have developed an algorithm for Q-Learning under Tsallis entropy regularization as summarized in Algorithm \ref{alg-1}.

\subsection{Bellman Residue of Entropy Regularized Q-Learning}
\label{sub-sec-br}

This subsection analyzes the Bellman residue to show that our Q-Learning algorithm will not suffer from any performance degradation despite of using approximated $\pi^*_{\alpha}$ in \eqref{equ-psa-prop1}. We will particularly consider one extreme circumstance when $\alpha\rightarrow 0$. With $\alpha$ approaching to 0, it is straightforward to verify that only the action that produces the highest Q-value in any state $s$, i.e. $a_1$, satisfies the condition in \eqref{equ-ss-approx-cond}. Therefore $c(s)=\alpha \left( \frac{Q^{\pi^*_{\alpha}}(s,a_1)}{\alpha}-\frac{q}{q-1}\right)$. According to \eqref{equ-psa-prop1}, $\pi^*_{\alpha}(s,a_1)\propto \delta(\frac{q}{q-1})>0$. On the other hand, for any action $a\neq a_1$,
\begin{equation}
\begin{split}
&\lim_{\alpha\rightarrow 0} \sqrt[q-1]{\delta\left( \frac{Q^{\pi^*_{\alpha}}(s,a)}{\alpha}-\frac{c(s)}{\alpha} \right)} \\
& = \lim_{\alpha\rightarrow 0} \sqrt[q-1]{\delta\left( \frac{Q^{\pi^*_{\alpha}}(s,a)-Q^{\pi^*_{\alpha}}(s,a_1)}{\alpha}+\frac{q}{q-1}\right)} \\
& = 0
\end{split}
\end{equation}

\noindent
We hence can conclude that, when $\alpha\rightarrow 0$, $\pi^*_{\alpha}(s,a_1)\rightarrow 1$. Consequently, $\pi^*_{\alpha}$ degenerates to the value-maximization policy. Based on this understanding, we can further prove Theorem \ref{theo-1} below (see Appendix for proof).
\begin{theorem}
For the Q-learning problem under Tsallis entropy regularization, suppose that $\pi^*_{\alpha}$ is the approximated optimal policy for the problem as defined in \eqref{equ-psa-prop1} and $\alpha$ is the corresponding regularization coefficient, then for any state $s\in\mathbb{S}$ and any action $a\in\mathbb{A}$, the Bellman residue $|\mathcal{T}^*Q^{\pi^*_{\alpha}}- Q^{\pi^*_{\alpha}}|$ satisfies the following property
$$
\lim_{\alpha\rightarrow 0} \left| \mathcal{T}^*Q^{\pi^*_{\alpha}}(s,a) - Q^{\pi^*_{\alpha}} (s,a) \right|=0
$$
\label{theo-1}
\end{theorem}

Because the Bellman residue converges to 0 with decreasing $\alpha$, when $\alpha$ is sufficiently small, the performance of Algorithm \ref{alg-1} is as good as DQN and Double DQN. For this reason, Algorithm \ref{alg-1} employs a linear schedule to constantly decrease $\alpha$, thereby gradually reducing entropy-induced exploration till $\alpha=0$ and learning converges.

\section{Bootstrapped Q-Ensemble under Tsallis Entropy Regularization}
\label{sec-deepexplor}

Using Algorithm \ref{alg-1} alone is insufficient to realize deep and effective exploration. Following the \emph{deep exploration principle} studied in \cite{osband2014,osband2017}, we expand an MDP $\mathcal{M}_q$ in this paper with the Tsallis entropy regularizer in \eqref{equ-tsa-entr} under the settings of $1<q<q_{max}<\infty$. In other words, the immediate reward of performing any action in state $s$ by following policy $\pi$ is extended with a new term that depends on Tsallis entropy of $\pi$ in state $s$. Given the experiences obtained so far by an RL agent through direct interactions with its learning environment, denoted as $\mathcal{B}$, a posterior model over $\mathcal{M}_q$ can be established in theory and represented as $P(\mathcal{M}_q|\mathcal{B})$.

Deep exploration requires an RL agent to randomly sample one MDP $\mathcal{M}_q$ from $P(\mathcal{M}_q|\mathcal{B})$ and subsequently utilize $Q^{\pi^*_{\alpha,\mathcal{M}_q}}$ and $\pi^*_{\alpha,\mathcal{M}_q}$ to control future interactions with its learning environment in the next problem episode. Each problem episode starts from an initial state $s_0$ and ends whenever a final state is reached. $Q^{\pi^*_{\alpha,\mathcal{M}_q}}$ and $\pi^*_{\alpha,\mathcal{M}_q}$ stand respectively for the optimal Q-function and optimal policy with respect to an MDP $\mathcal{M}_q$ under the specific settings of $q$ and $\alpha$. Apparently this deep exploration method has the aim of optimizing the posterior learning performance of an RL agent based on its past experiences, as described below
\begin{equation}
\begin{split}
J^*(\mathcal{B})= & \E_{\mathcal{M}_q\sim P(\mathcal{M}_q|\mathcal{B})}\\
& \E_{(s,a)\sim\pi^*_{\alpha,\mathcal{M}_q}} \left( r(s,a)+\alpha H_q^{\pi^*_{\alpha,\mathcal{M}_q}}(s) \right)
\end{split}
\label{equ-de-goal}
\end{equation}

On large-scale RL problems it is difficult to keep track of $P(\mathcal{M}_q|\mathcal{B})$ as well as to determine the optimal policy for every possible $\mathcal{M}_q$. Inspired by \cite{osband20161}, we decide to efficiently approximate the deep exploration process through bootstrapping. This is implemented in the BQETR algorithm (see Algorithm \ref{alg-2}) by maintaining an ensemble of entropy regularized Q-networks. All Q-networks share the same regularization coefficient $\alpha$. Meanwhile, different Q-networks follow different settings of $q$ so as to enhance the diversity of the ensemble, which is essential for effective deep exploration.

Apparently, no change to $q$ is required for any Q-network in the ensemble during RL. Since agent's past experiences $\mathcal{B}$ will not affect the posterior distribution over $q$, each Q-network in the ensemble can be sampled equally likely for the next episode of deep exploration. This simple technique enables us to seamlessly integrate entropy-induced exploration with bootstrap-induced exploration and lays the foundation of the BQETR algorithm. Because $\alpha$ is decremented each time by a very small step $\Delta_{\alpha}$ in Algorithm \ref{alg-1} and Algorithm \ref{alg-2}, its change will not affect the effectiveness of the bootstrapping mechanism.

\begin{algorithm}[!ht]
 \begin{algorithmic}[1]
 \item {\bf Input}: an ensemble of $K$ Q-networks $\{Q_k\}_{k=1}^K$, a list of $q$ values $\{q_k\}_{k=1}^K$ for the Tsallis entropy regularizers, $\alpha_0$ for the initial regularization coefficient, a replay buffer $\mathcal{B}$ that stores past state-transition samples for training, and a masking distribution $M$.
 \item {\bf for} each problem episode {\bf do}:
 \State \ \ \ \ Choose the $i$-th Q-network in $\{Q_k\}_{k=1}^K$ randomly
 \item \ \ \ \ Obtain initial state $s_0$ from environment
 \item \ \ \ \ {\bf for} $t=1,\ldots$ until end of episode {\bf do}:
 \State \ \ \ \ \ \ \ \ Use $Q_i$, $q_i$ and $\alpha$ to sample action $a_t$

 \hspace{0.0cm} according to \eqref{equ-psa-prop1}

 \item \ \ \ \ \ \ \ \ Perform $a_t$
 \item \ \ \ \ \ \ \ \ Sample bootstrap mask $m_t\sim M$
 \item \ \ \ \ \ \ \ \ Add $(s_t,a_t,s_{t+1},r_t,m_t)$ to $\mathcal{B}$
 \item \ \ \ \ \ \ \ \ {\bf if} learning interval is reached {\bf do}:
 \item \ \ \ \ \ \ \ \ \ \ \ \ Follow Algorithm \ref{alg-1} to train all $K$

 \ \ \ \ \ Q-networks.
 \item \ \ \ \ \ \ \ \ \ \ \ \ Reduce $\alpha$ linearly by $\Delta_{\alpha}$ until 0
 \end{algorithmic}
\caption{The Bootstrapped Q-Ensemble under Tsallis Entropy Regularization (BQETR) Algorithm}
\label{alg-2}
\end{algorithm}

\begin{figure*}[t]
      \centering
      \begin{minipage}[t]{0.315\textwidth}
        \centering
        \includegraphics[width=\textwidth]{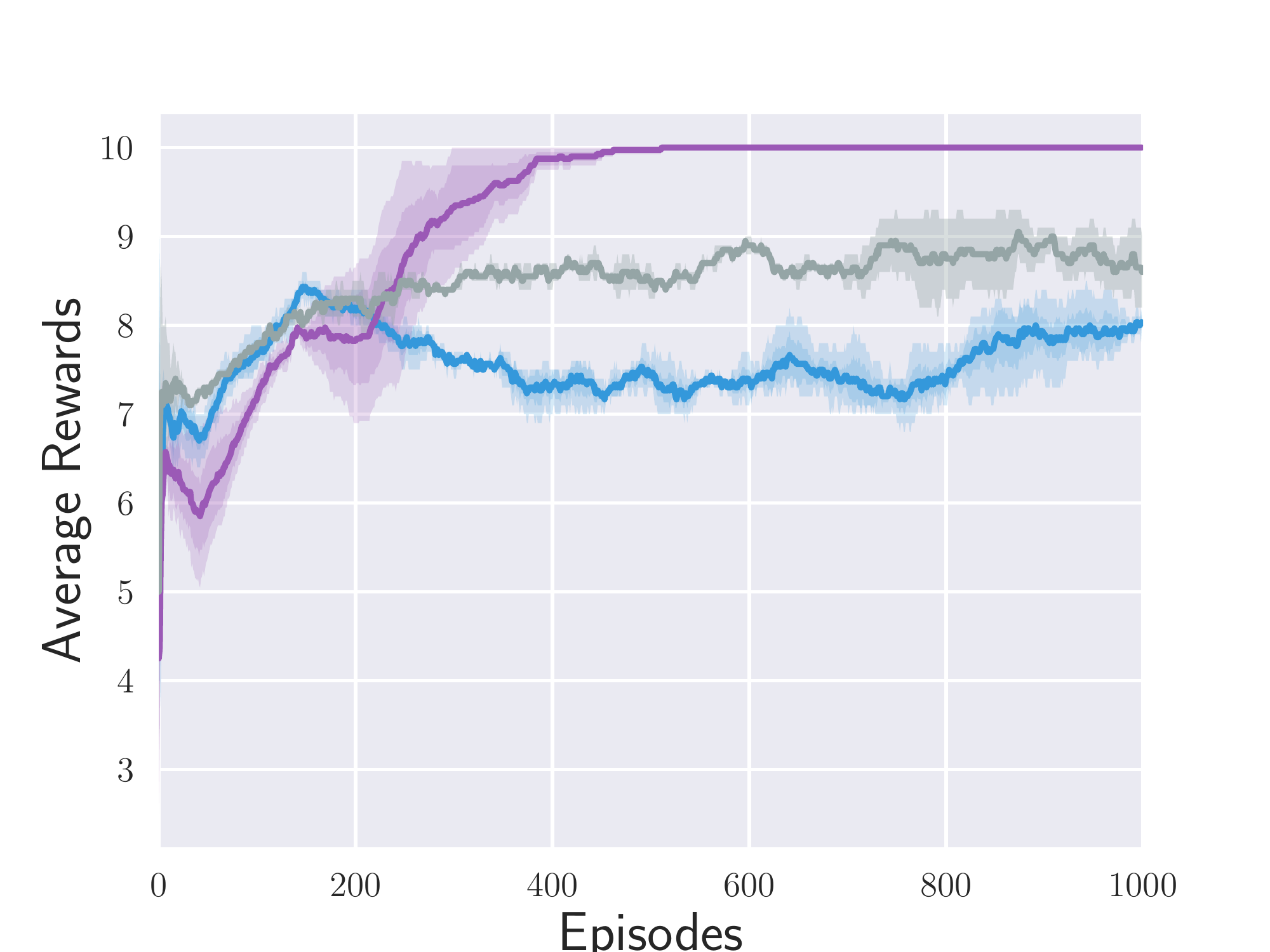}
        \subcaption{Bowling}
      \end{minipage}%
      \begin{minipage}[t]{0.315\textwidth}
        \centering
        \includegraphics[width=\textwidth]{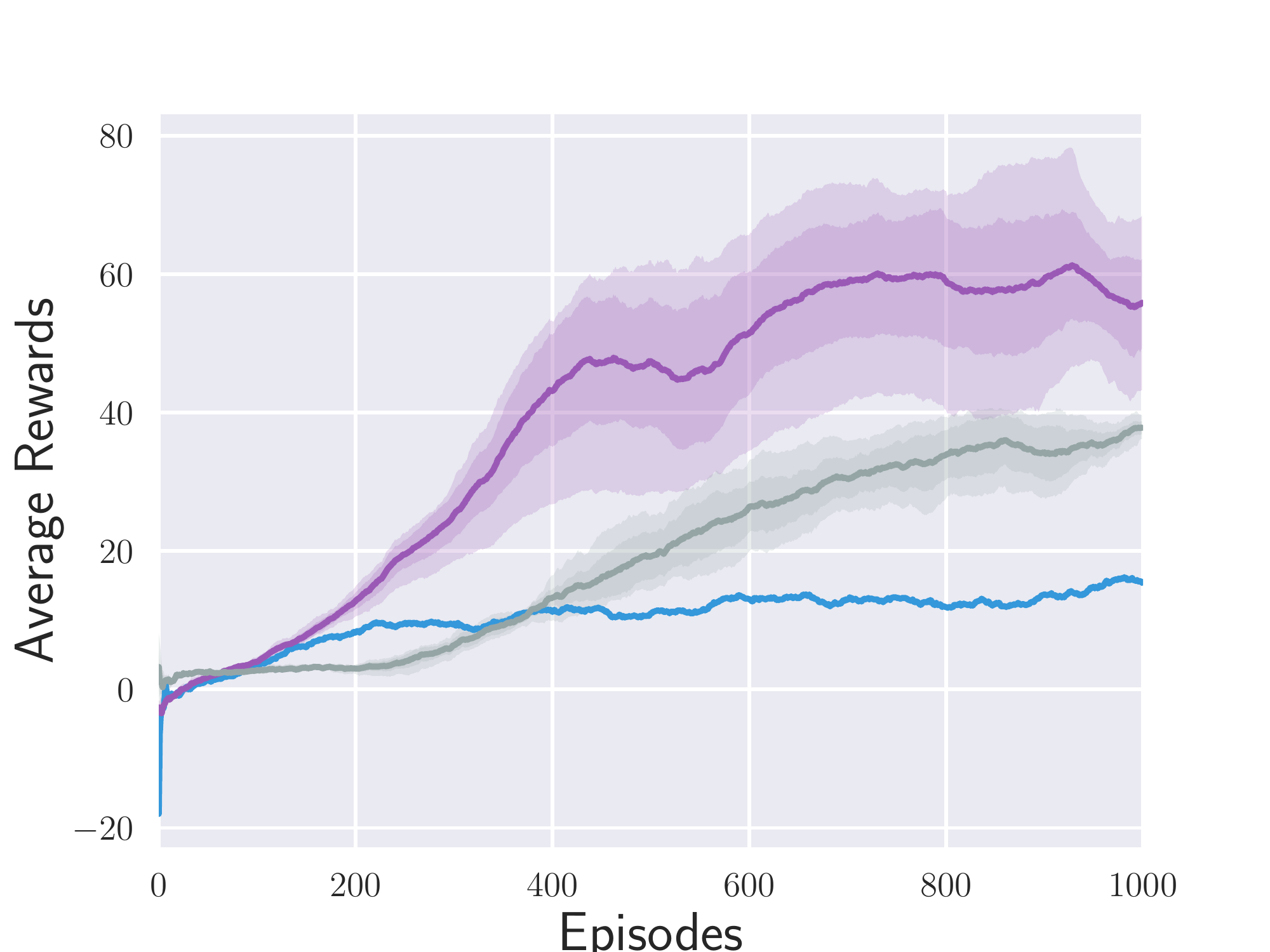}
        \subcaption{Boxing}
      \end{minipage}
      \begin{minipage}[t]{0.315\textwidth}
        \centering
        \includegraphics[width=\textwidth]{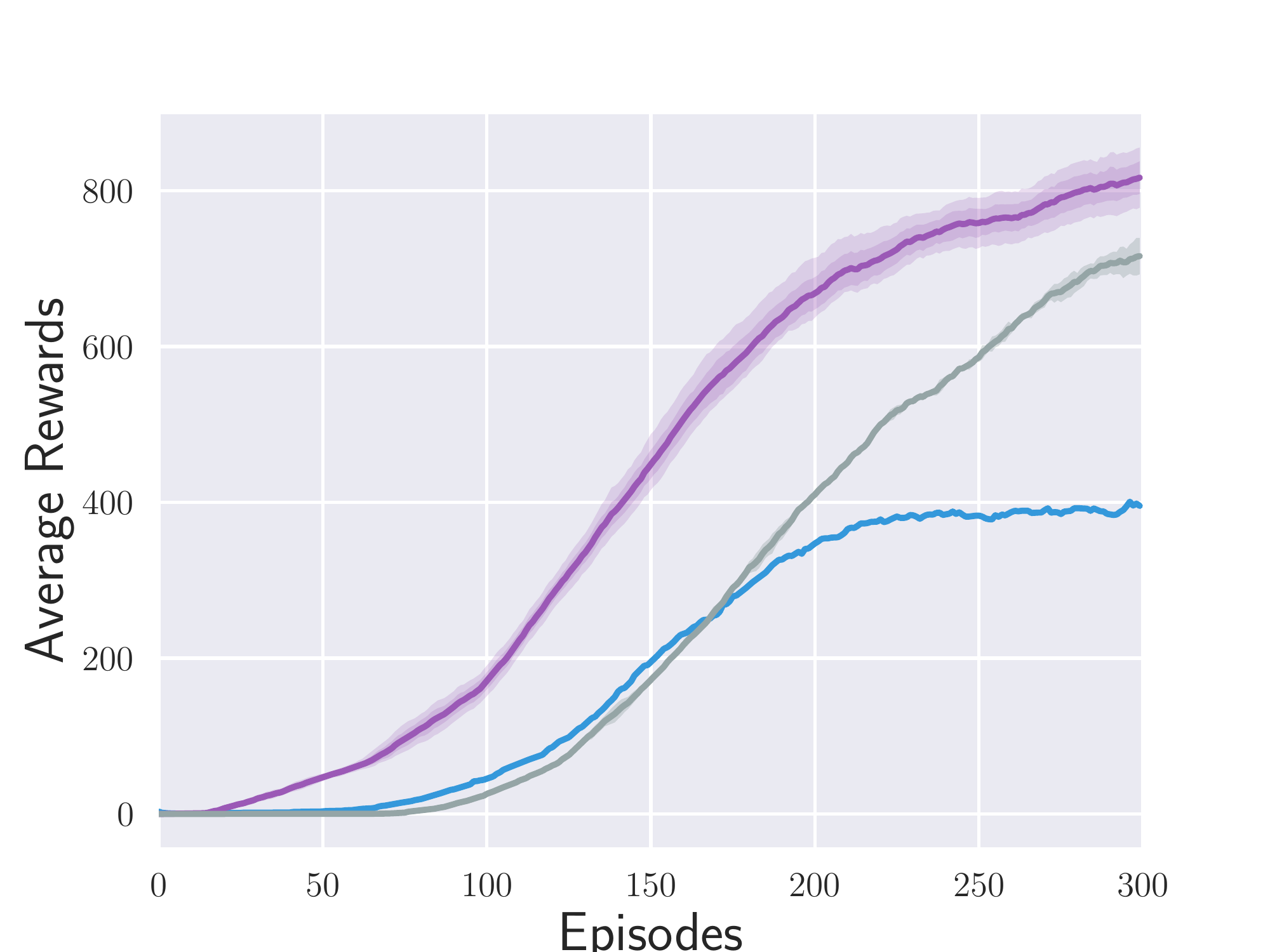}
        \subcaption{Enduro}
      \end{minipage}
      \\
      \begin{minipage}[t]{0.315\textwidth}
        \centering
        \includegraphics[width=\textwidth]{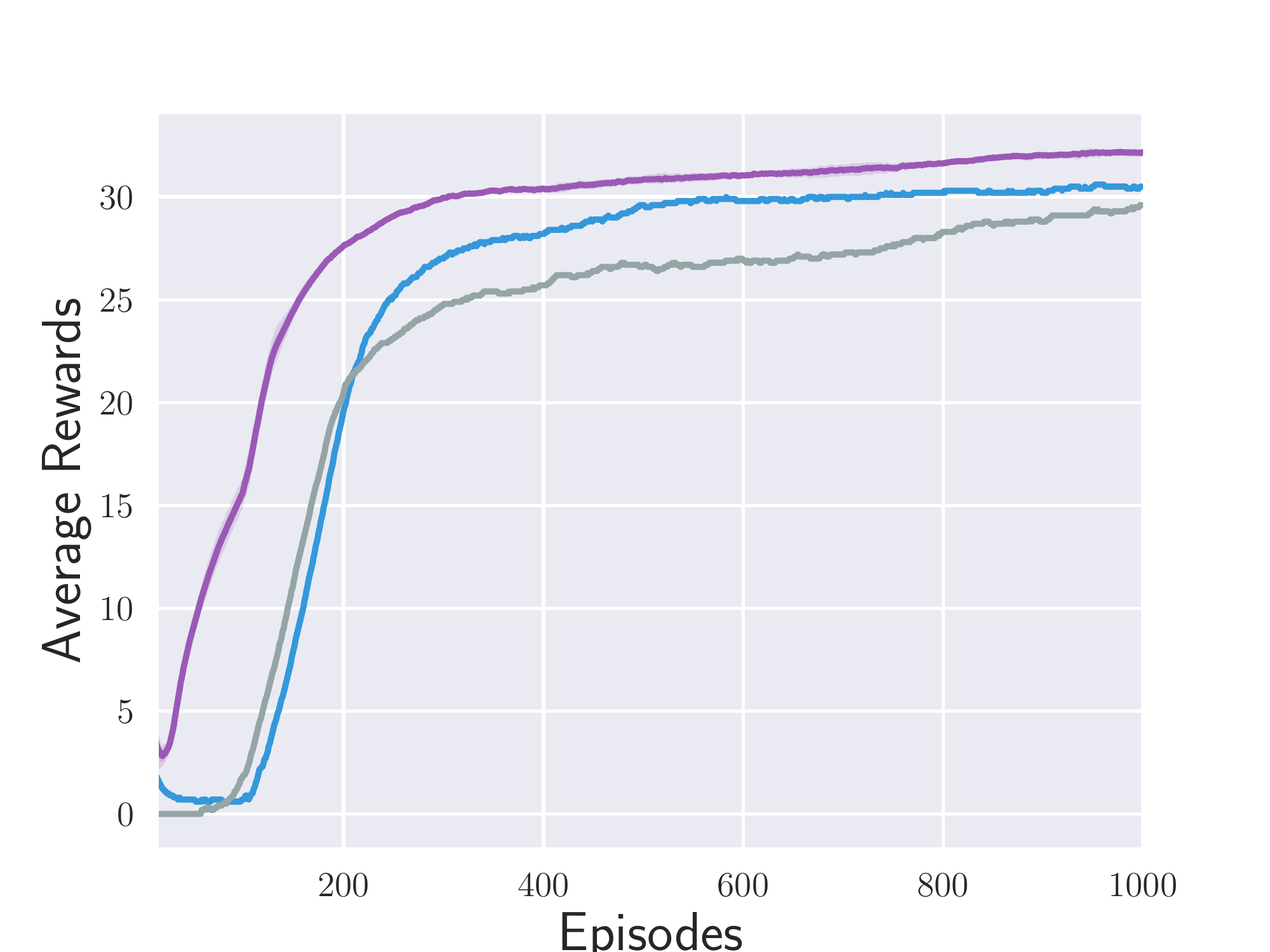}
        \subcaption{Freeway}
      \end{minipage}
      \begin{minipage}[t]{0.315\textwidth}
        \centering
        \includegraphics[width=\textwidth]{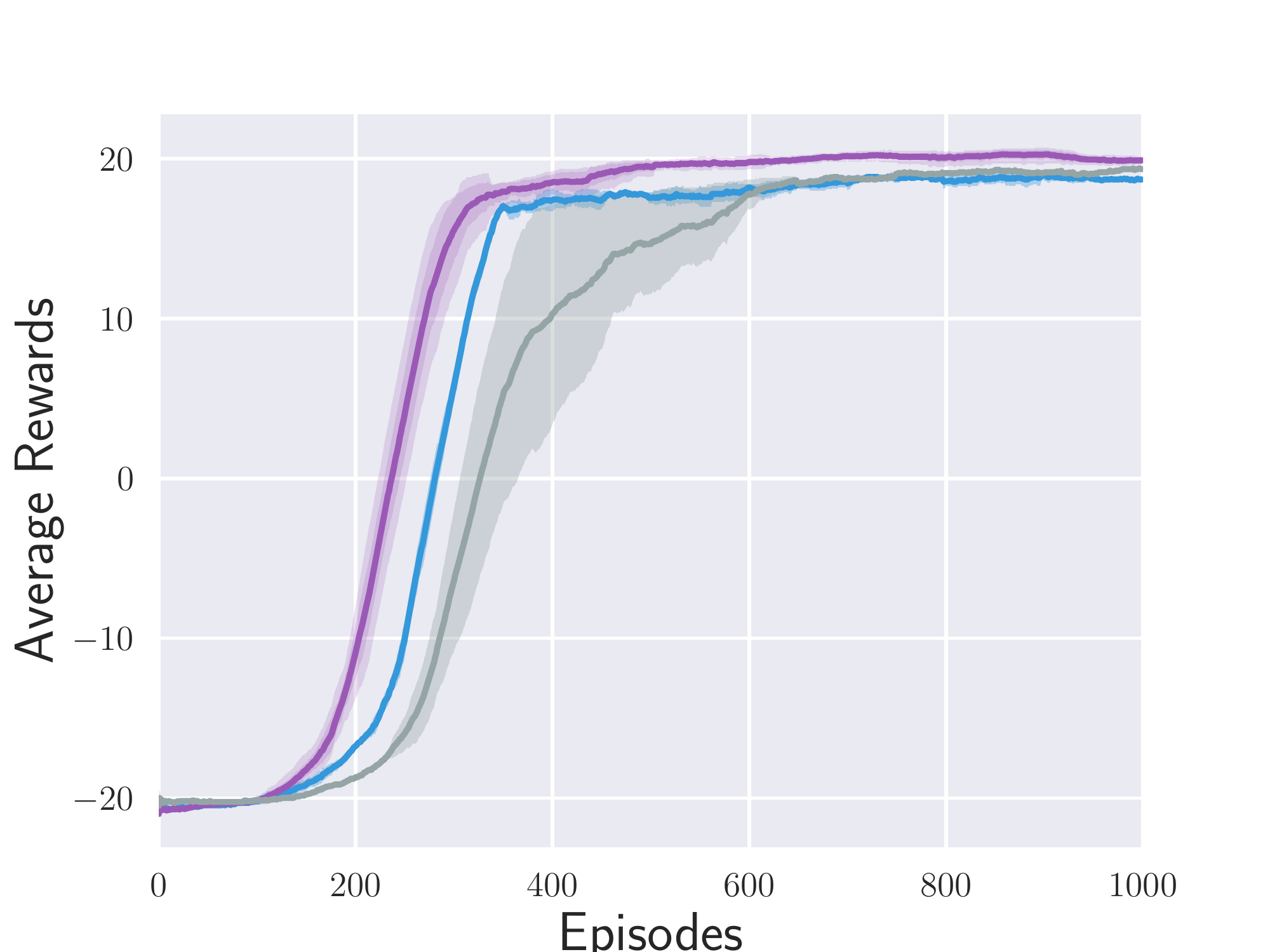}
        \subcaption{Pong}
      \end{minipage}
      \begin{minipage}[t]{0.315\textwidth}
        \centering
        \includegraphics[width=0.6\textwidth]{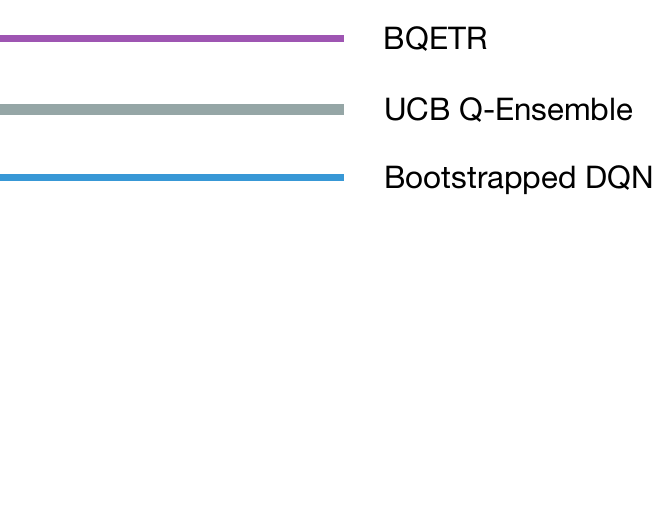}
        % \subcaption{Legend}
      \end{minipage}
      \caption{Average total return per episode obtained by BQETR, Bootstrapped DQN and UCB Q-Ensemble on five Atari game playing tasks, including Bowling, Boxing, Enduro, Freeway, and Pong.}
      \label{fig-lea-eff}
  \end{figure*}

\section{Experiment}
\label{sec-exp}

In this section, the learning performance of BQETR is compared to the performance achievable through Bootstrapped DQN and UCB Q-Ensemble on commonly studied Atari game playing tasks \cite{van2016,mnih2015,schulman2015,osband20161,chen2017ucb}. Based on the performance results, the sample complexity of the three algorithms is further analyzed to demonstrate that BQETR can improve sample efficiency through deep and effective exploration of its learning environment.

In this paper we consider specifically five video games simulated by the Arcade Learning Environment \cite{Bellemare2015} as benchmark problems, including Bowling, Boxing, Enduro, Freeway and Pong. These problems require an RL agent to handle high-dimensional state spaces (i.e. an agent must be able to process direct video input provided by the games) and are highly difficult to solve, even for expert human game players. As a result, to achieve reasonable learning performance, an RL agent must play numerous rounds of each benchmark game. Hence they are suitable problems to reveal the difference in sample efficiency upon using various exploration methods for DRL.

We implement all algorithms in the experiments based on the high-quality implementation of Double DQN provided by OpenAI baselines \cite{baselines}. We also closely follow the parametric settings of Boostrapped DQN and UCB Q-Ensemble presented in \cite{osband20161,chen2017ucb}. Meanwhile, for a fair comparison, we use identical settings for common parameters shared between BQETR and Bootstrapped DQN. BQETR also introduces two additional parameters, i.e. the initial value for the entropy regularization coefficient $\alpha_0$ and the \emph{entropic index} $q$ for Tsallis entropy. Without spending substantial efforts in fine-tuning these parameters, $\alpha_0$ is set to 0.5 (other settings ranging from 1.0 to 0.1 do not seem to produce noticeable difference in performance). After each learning interval, $\alpha$ will be decremented by $0.5\times 10^{-5}$ till 0. Since the Q-ensemble maintained by all algorithms contains 10 individual Q-networks. The values for $q$ in BQETR have been set to $1.5, 1.6, \ldots, 2.4$ respectively for each Q-network. Moreover, every pixel input to a Q-network is obtained by averaging the same pixel over four consecutive frames of the game video. On each game playing task, we have run every algorithm for only 3M frames by using commodity desktop computers (no GPUs). This enables us to examine the effectiveness of all algorithms under limited computation resources and sample budget.

\subsection{Results on Learning Effectiveness}
\label{sub-exp-lea-eff}

Figure \ref{fig-lea-eff} depicts the learning performance (i.e., average total return per episode) of the three algorithms in our experiments. To cover at least 3M frames, the performance across 1000 learning episodes have been presented in the figure, except for Enduro. This is because each episode in Enduro includes more frames. Therefore 3M frames have been reached after just playing 300 episodes of Enduro.

As evidenced in Figure \ref{fig-lea-eff}, BQETR outperformed Bootstrapped DQN and UCB Q-Ensemble on all benchmark problems.
Particularly on Bowling, Boxing and Enduro, BQETR achieved significantly higher performance than competing algorithms. In the meantime, BQETR also managed to solve Freeway and Pong clearly faster than other algorithms. Based on the experiment results, we believe that BQETR is an effective algorithm for DRL thanks to its integrated use of both entropy-induced and bootstrap-induced exploration techniques.

\subsection{Results on Sample Efficiency}
\label{sub-exp-sam-eff}

To analyze sample efficiency, we adopt the performance metrics introduced in \cite{schulman20171}. Particularly, the fast learning metric in Table \ref{table-sam-eff} calculates the average performance across all learning episodes and the final performance metric calculates the average performance obtained in the last 10 episodes. Both metrics in Table \ref{table-sam-eff} clearly show that BQETR is not only effective in terms of final performance but also significantly more sample efficient than both Bootstrapped DQN and UCB Q-Ensemble.

\begin{table}[!ht]
\center
\scalebox{0.7}{
\begin{tabular}{c|c|lllll}
\hline
\textbf{Scoring Metric}            & \textbf{Algorithms} & \textbf{Bowling}           & \textbf{Boxing}             & \textbf{Enduro}              & \textbf{Freeway}           & \textbf{Pong}              \\ \hline
\multirow{3}{*}{Fast Learning}     & BQETR               & \textbf{9.36 $\rpm$ 0.47}  & \textbf{60.61 $\rpm$ 17.71} & \textbf{486.13 $\rpm$ 34.56} & \textbf{29.33 $\rpm$ 0.63} & \textbf{12.76 $\rpm$ 0.41} \\
                                   & Bootstrapped DQN    & 7.63 $\rpm$ 0.55           & 19.80 $\rpm$ 1.00           & 279.98 $\rpm$ 10.00           & 26.19 $\rpm$ 0.30          & 11.47 $\rpm$ 0.24          \\
                                   & UCB Q-Ensemble      & 8.59 $\rpm$ 0.41           & 38.11 $\rpm$ 6.36          & 390.68 $\rpm$ 24.10          & 25.08 $\rpm$ 0.33          & 10.72 $\rpm$ 1.32          \\ \hline
\multirow{3}{*}{Final Performance} & BQETR               & \textbf{10.00 $\rpm$ 0.02} & \textbf{57.41 $\rpm$ 7.11} & \textbf{812.10 $\rpm$ 36.39} & \textbf{32.22 $\rpm$ 0.28} & \textbf{20.74 $\rpm$ 0.38} \\
                                   & Bootstrapped DQN    & 7.97 $\rpm$ 0.73           & 41.50 $\rpm$ 1.00           & 374.23 $\rpm$ 5.00           & 30.86 $\rpm$ 0.10          & 19.23 $\rpm$ 0.18          \\
                                   & UCB Q-Ensemble      & 9.20 $\rpm$ 0.09           & 47.96 $\rpm$ 2.34          & 725.95 $\rpm$ 30.50         & 31.00 $\rpm$ 0.10          & 19.34 $\rpm$ 0.30          
\end{tabular}}
\caption{Scoring metrics of fast learning and final performance obtained by BQETR, Bootstrapped DQN and UCB Q-Ensemble on five Atari games.}
\label{table-sam-eff}
\end{table}

\section{Conclusions}
\label{sec-con}

In this paper we studied entropy-induced environment exploration via deep Q-learning under general Tsallis entropy regularization. Through this study, we developed the first time in literature new approximation techniques to address entropy regularized RL problems. Bellman residue analysis subsequently showed that our approximation techniques will not affect the final performance achievable through Q-learning. Driven by the goal for deep exploration, we have further developed a bootstrapped Q-learning algorithm involving an ensemble of Q-networks. Every Q-network is controlled by a Tsallis entropy regularizer under different settings of $q$ so as to achieve high ensemble diversity and effective deep exploration.

Looking into the future, it is interesting to explore the possibilities of extending our Q-learning algorithm to tackle RL problems with high-dimensional and continuous action spaces. Meanwhile, it is also interesting to explore the benefits of our Q-learning algorithm on more problem domains. Due to limited computation resources that are available to this research, we cannot conduct large-scale experimental studies in this paper. However our experiment results have clearly shown that our new algorithm is both effective and sample efficient.

\section*{Appendix}

This appendix presents proof of Theorem \ref{theo-1}, i.e. $\|\mathcal{T}^*Q^{\pi^*_{\alpha}}- Q^{\pi^*_{\alpha}}\|\rightarrow 0$ by decreasing the regularization coefficient $\alpha$ all the way to 0. Specifically, for any state-action pair $(s,a)$, we can derive the following inequalities.
\begin{equation}
\begin{split}
0 & \leq \left| \mathcal{T}^*Q^{\pi^*_{\alpha}}(s,a) - \mathcal{T}^{\alpha} Q^{\pi^*_{\alpha}} (s,a) \right| \\
& \leq \E_{s'\sim P(s,a,s')} \left| \max_b Q^{\pi^*_{\alpha}}(s',b)-\sum_b \pi^*_{\alpha}(s',b) Q^{\pi^*_{\alpha}}(s',b) \right| \\
& \leq \E_{s'\sim P(s,a,s')} \left( \sum_b |\mathbb{I}_{b=b_1}^{s'}- \pi^*_{\alpha}(s',b) | \cdot |Q^{\pi^*_{\alpha}}(s',b)| \right) \\
& \leq \E_{s'\sim P(s,a,s')} \left( \sum_b |\mathbb{I}_{b=b_1}^{s'}- \pi^*_{\alpha}(s',b) | \sum_b |Q^{\pi^*_{\alpha}}(s',b)| \right)
\end{split}
\label{equ-br-ine1}
\end{equation}

\noindent
where $\mathbb{I}_{b=b_1}^{s'}$ refers to the policy that selects action $b_1$ in state $s'$ with probability 1 and $b_1$ is the action with the highest Q-value in state $s'$. Assume without loss of generality that the absolute Q-value with respect to any state and any action can never exceed $\bar{Q}$. Then, from \eqref{equ-br-ine1}, we have
\begin{equation}
\begin{split}
&\left| \mathcal{T}^*Q^{\pi^*_{\alpha}}(s,a) - \mathcal{T}^{\alpha} Q^{\pi^*_{\alpha}} (s,a) \right| \\
& \leq \|\mathbb{A}\|\cdot \bar{Q}\cdot \E_{s'\sim P(s,a,s')} \left( 2 D_{TV}\left( \mathbb{I}_{\cdot=b_1}^{s'} \| \pi^*_{\alpha}(s',\cdot) \right) \right)
\end{split}
\label{equ-br-ine2}
\end{equation}

\noindent
with $D_{TV}(x\|y)=\frac{1}{2}\sum_i |x_i-y_i|$ representing the \emph{total variation divergence} in between any two discrete probability distributions $x$ and $y$. Due to the fact that $D_{TV}(x|y)^2\leq D_{KL}(x\|y)$ where $D_{KL}$ is the standard \emph{KL divergence} \cite{pollard2000}, we can further obtain the inequality below from \eqref{equ-br-ine2},
\begin{equation}
\begin{split}
&\left| \mathcal{T}^*Q^{\pi^*_{\alpha}}(s,a) - \mathcal{T}^{\alpha} Q^{\pi^*_{\alpha}} (s,a) \right| \\
& \leq 2 \|\mathbb{A}\|\cdot \bar{Q} \cdot \E_{s'\sim P(s,a,s')}\sqrt{D_{KL}\left(\mathbb{I}_{\cdot=b_1}^{s'} \| \pi^*_{\alpha}(s',\cdot)  \right)} \\
& = 2 \|\mathbb{A}\|\cdot \bar{Q} \cdot \E_{s'\sim P(s,a,s')} \sqrt{-\log \pi^*_{\alpha}(s',b_1)}
\end{split}
\label{equ-br-ine3}
\end{equation}

\noindent
Consequently,
\begin{equation}
\begin{split}
0 &\leq \lim_{\alpha\rightarrow 0} \left| \mathcal{T}^*Q^{\pi^*_{\alpha}}(s,a) - \mathcal{T}^{\alpha} Q^{\pi^*_{\alpha}} (s,a) \right| \\
 & \leq 2 \|\mathbb{A}\|\cdot \bar{Q} \cdot \E_{s'\sim P(s,a,s')} \lim_{\alpha\rightarrow 0} \sqrt{-\log \pi^*_{\alpha}(s',b_1)} \\
 & = 0
\end{split}
\label{equ-br-ine4}
\end{equation}

\noindent
Using \eqref{equ-br-ine4}, it can be further shown that
\begin{equation}
\begin{split}
0 \leq &\lim_{\alpha\rightarrow 0} \left| \mathcal{T}^*Q^{\pi^*_{\alpha}}(s,a) - Q^{\pi^*_{\alpha}} (s,a) \right| \\
\leq & \lim_{\alpha\rightarrow 0} \left| \mathcal{T}^*Q^{\pi^*_{\alpha}}(s,a) - \mathcal{T}^{\alpha} Q^{\pi^*_{\alpha}} (s,a) \right| \\ & + \lim_{\alpha\rightarrow 0} \left| \mathcal{T}^{\alpha} Q^{\pi^*_{\alpha}}(s,a) - Q^{\pi^*_{\alpha}} (s,a) \right| \\
= & 0
\end{split}
\label{equ-br-ine5}
\end{equation}

\noindent
Notice that when $\alpha\rightarrow 0$, the error involved in approximating $\pi^*_{\alpha}$ in \eqref{equ-pi-sol} is negligible since the action that produces the highest Q-value will be selected with probability 1. As a result, $\mathcal{T}^{\alpha} Q^{\pi^*_{\alpha}}(s,a) - Q^{\pi^*_{\alpha}} (s,a) = 0$ for any state $s$ and action $a$.

% OBLIGATORY: use BIBTEX formatting!
\small{
\bibliographystyle{alpha}
\bibliography{citefile}
}
\normalsize

\end{document}